\documentclass{article}
\usepackage{myarxiv}

\usepackage{times}
\usepackage{soul}
\usepackage{url}
\usepackage[utf8]{inputenc}
\usepackage[small]{caption}
\usepackage{graphicx}
\usepackage{amsmath}
\usepackage{amsfonts}
\usepackage{makecell}
\usepackage{booktabs}
\usepackage{multirow}
\usepackage{algorithm}
\usepackage{algpseudocode}
\usepackage[tight,footnotesize]{subfigure}
\DeclareMathOperator*{\argmax}{argmax}

\title{Evolutionary Trigger Set Generation for DNN Black-Box Watermarking}

\author{
Jia Guo
\and
Miodrag Potkonjak
\affiliations
Computer Science Department, UCLA\\
\emails
\{jia, miodrag\}@cs.ucla.edu.,
}

\begin{document}

\maketitle

\begin{abstract}
The commercialization of deep learning creates a compelling need for intellectual property (IP) protection. Deep neural network (DNN) watermarking has been proposed as a promising tool to help model owners prove ownership and fight piracy. A popular approach of watermarking is to train a DNN to recognize images with certain \textit{trigger} patterns. In this paper, we propose a novel evolutionary algorithm-based method to generate and optimize trigger patterns. Our method brings siginificant reduction in false positive rates, leading to compelling proof of ownership. At the same time, it maintains the robustness of the watermark against attacks. We compare our method with prior art and demonstrate its effectiveness on popular models and datasets \footnote{\url{https://github.com/guojia-git/watermarking-cnn-classifiers}}.
\end{abstract}

\section{Introduction}

Since the success of large scale neural networks in the early 2010s, we have witnessed an explosive growth in the field. The popularity grew not only in academia but in the industry as well. Deep neural networks (DNNs) have become the de facto solution to many complex computer vision, speech recognition, and natural language processing problems. Popular as deep learning may be, building DDNs to solve real-world problems remains an arduous task. It requires a vast amount of high-quality labeled data and heavy use of computational resources and human expertise. It goes without saying that DNNs are invaluable technological assets that potentially has huge commercial impacts. Over the past few years, myriads of companies have joined the AI arms race. Just among AI startups, investment from the venture capital market reached a record high \$9.3 billion in 2018 \cite{ai-investment}. Among the companies, many provide range from commercial libraries for embedded systems, cloud machine learning APIs to building private corporate clouds for AI, spanning across industries like transportation, manufacturing, healthcare, finance, and consumer electronics.

\begin{figure}
    \centering
    \includegraphics[width=3.6in]{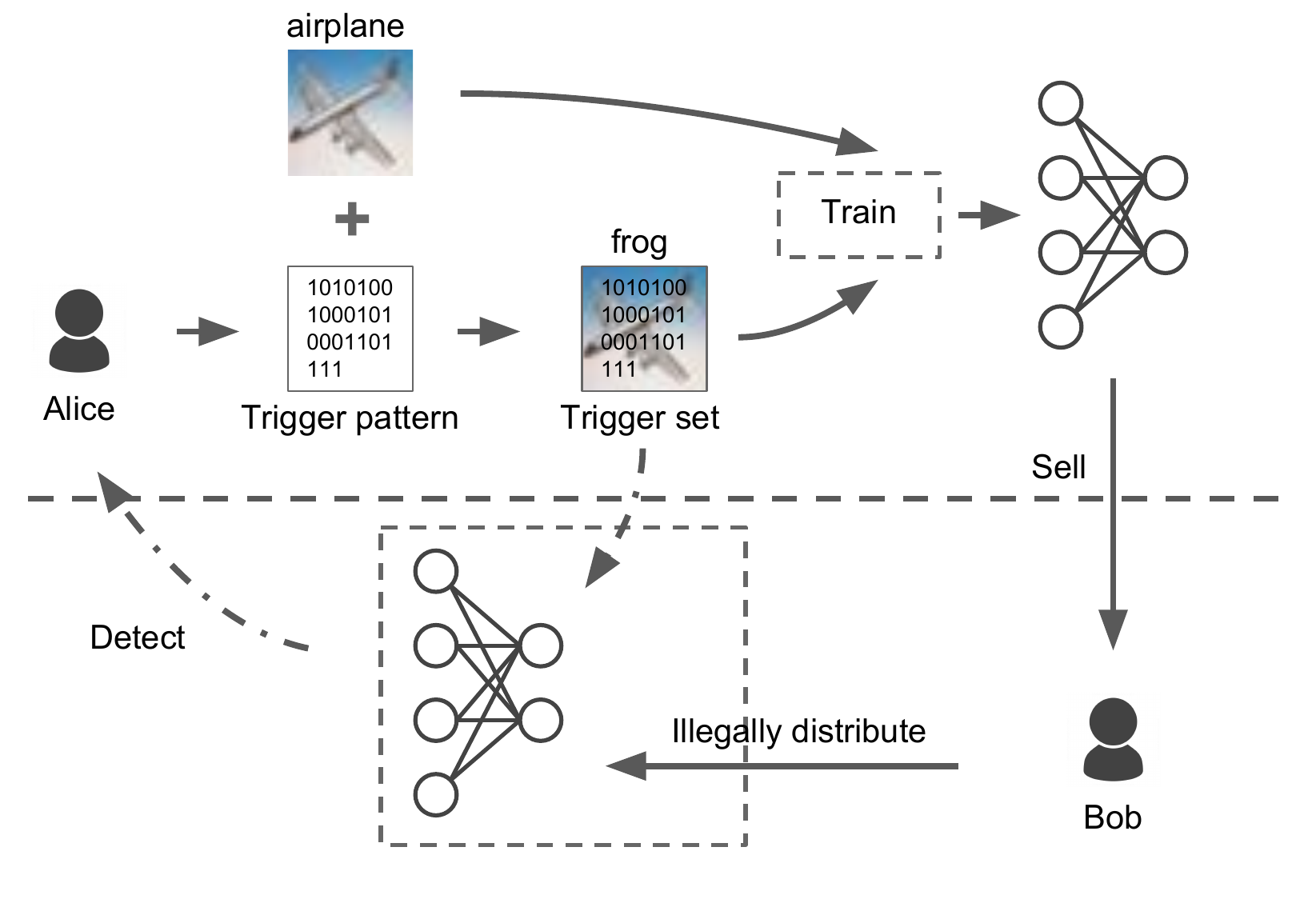}
    \caption{Workflow of the trigger pattern-based black-box DNN watermarking.}
    \label{fig:workflow}
\end{figure}

While the DNNs are fueling commercial successes in the AI market, a void in IP protection for DNN models may hinder the progress. When a model owner sells a service to a customer, she should have a reliable way to prevent the customer from illegally distributing or reselling it. To achieve that goal, the owner not only needs to \textit{identify} her own model when it is distributed, but also \textit{prove} the ownership to a trusted arbitrator. Recently, several researchers proposed watermarking as a viable solution to the IP protection problem in deep learning \cite{UchidaNSS17}\cite{abs-1711-01894}\cite{AdiBCPK18} \cite{abs-1804-00750}\cite{ZhangGJWSHM18}\cite{GuoP18}. Digital watermarking originally refers to the process of covertly embedding information in multimedia content. The concept has since been extended to cover software \cite{CollbergT99}, circuits \cite{KahngLMMMPTWW98} as well as DNNs. \textit{White-box} watermarking embeds the owner's information in the weights of a DNN. \textit{Black-box} watermarking, on the other hand, embeds the watermark in the input-output behavior of the model. The set of input used to trigger that behavior is called \textit{trigger set}. For the popular task of image classification, a common approach is to assign a random label to trigger images and train the model to classify accordingly. The non-triviality of ownership of a watermarked model is constructed on the extremely small probability for any other model to exhibit the same behavior \cite{AdiBCPK18}. Through detecting the watermark in a DNN model, the owner will be able to both identify and prove her ownership.

\begin{figure}
    \centering
    \subfigure[]{
        \includegraphics[width=.9in,height=.9in]{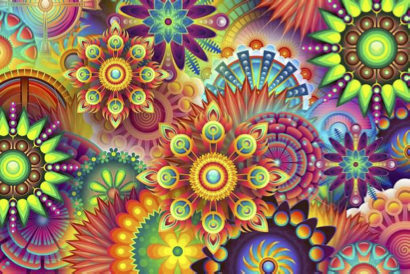}
        \label{fig:other_1}
    }  
    %\hspace{-0.em}
    \subfigure[]{
        \includegraphics[width=.9in]{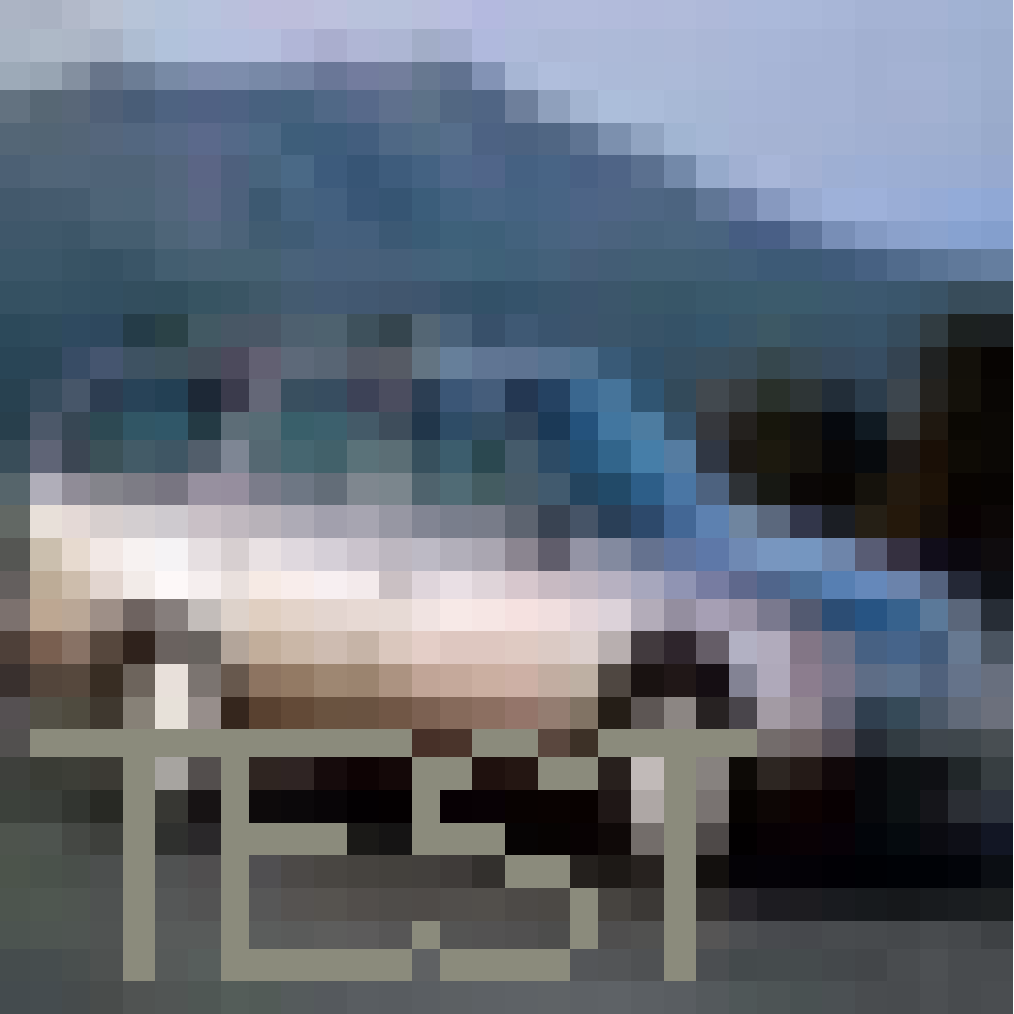}
        \label{fig:other_2}
    }
    %1\hspace{-0.5em}    
    \subfigure[]{
        \includegraphics[width=.9in]{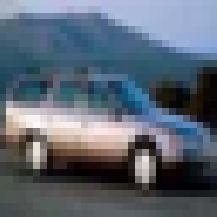}
        \label{fig:other_3}
    }
    \caption{Examples of trigger inputs. (a) a random out-of-distrbution image \protect\cite{AdiBCPK18}, (b) a regular image with a logo \protect\cite{ZhangGJWSHM18}, (c) a regular image with a color-coded key not perceptible by the eye \protect\cite{GuoP18}. }
    \label{fig:other}
\end{figure}

Based on the characteristics of their trigger sets, existing black-box watermarking methods can be split into two categories. The first category of methods curates a finite set of special trigger images. The special images can be completely random \cite{AdiBCPK18}, samples derived from unused hidden space \cite{abs-1804-00750}, or adversarial examples \cite{abs-1711-01894}. Another category of methods maintain \textit{trigger patterns} and add them to natural input images to create trigger sets. The trigger patterns are usually meaningful patterns that can serve as a proof of the owner's identity, such as the logos \cite{ZhangGJWSHM18} and color-coded keys \cite{GuoP18}. Figure \ref{fig:workflow} describes the workflow of the method. Some sample trigger images are shown in Figure \ref{fig:other}. 

The motivation for designing the trigger sets are different between those two categories of methods. The first is focused on the functionality of the model. They aim to create trigger sets such that watermarking is as orthogonal to the normal functionality of the model as possible. The second, on the other hand, is more focused on the watermarking extraction procedure. Associating the owner's identity with the trigger set makes detection and proof of ownership much more straightforward. The evident drawback of the first category is the difficulty to establish a connection between the trigger set and the owner. To solve that problem, researchers went as fars as to use complex cryptographic tools \cite{AdiBCPK18}. Further, the limited size of the trigger set weakens the proof of ownership. The drawback of the second category lies in an inevitable trade-off between the robustness of the watermark and potential of false positive watermark detection. If a trigger pattern is too prominent, then it risks triggering false positives in other neural networks. On the other, if a pattern is too inconspicuous, it may be easily removed during fine-tune attacks.

In this paper, we aim to bridge the gap between the different trigger set generation methods. We propose a differential evolution-based framework to determine how \textit{any} given trigger pattern should be added to the image such that false positive detections are reduced while the robustness of the watermark is maintained. With our framework, trigger pattern-based watermarking adds the model functionality to its equation, while still keeping ownership proofs simple. The contribution of our paper are as follows:
\begin{itemize}
\item We proposed an evolutionary algorithm-based framework to optimize trigger patterns in order to facilitate robust and low-false-positive black-box watermarking 
\item We surveyed and compared existing trigger set generation methods and presented our analysis
\item We implemented our method with popular DNN models and datasets and evaluate its performance
\end{itemize}

The rest of the paper is organized as follows: Section \ref{problem} describes the watermarking problem in more details and defines the problem. Section \ref{method} presents our algorithm. Section \ref{evaluation} evaluates the performance of the proposed algorithm.

%\textit{Private watermarking systems} refer to systems whose watermarks are not accessible by the public \cite{cox2007digital}. Often there is a need for a trusted third party who extracts the watermark and verifies the ownership based on the watermark. The said third party should be honest and should not release the private watermark to the public \cite{AdiBCPK18}.

\section{Preliminaries}
\label{problem}
In this section, we will first introduce the background of watermarking. Then we will define our problem. Similar to most of the security-related literature, we use the Alice / Bob narrative to describe the scenario. Alice will be the model owner. Bob will be the customer who buys the model from Alice, and also the malicious attacker who tries to infringe on Alice's IP rights.

\subsection{DNN Watermarking}

\begin{table}[ht]
\begin{center}
\caption{Criteria for evaluating DNN watermarks.}
\label{tab:criteria}
\begin{tabular}{cl} \hline
\textbf{Criterion} & \textbf{Explanation} \\ \hline
Effectiveness & \makecell[l]{The watermarking method can prove \\ ownership in different DNNs and datasets} \\ \hline
Fidelity & \makecell[l]{The watermark does not significantly affect \\ the performance of the model} \\ \hline
\makecell[c]{False Positive\\Rate} & \makecell[l]{The watermark detection method will not \\ be triggered when there is no watermark} \\ \hline
\makecell[c]{Robustness} & \makecell[l]{The watermark is robust against attacks} \\ \hline
\end{tabular}
\end{center}
\end{table}

We define DNN watermarking as the process of covertly embedding information in the DNN in order to verify and prove an owner's ownership. We focus on black-box watermarking for image classification, which achieves the aforementioned goal by embedding special input-output patterns in DNNs. To embed the watermark, Alice will train a DNN with both the regular dataset and the trigger set with specific output labels. To detect the watermark, Alice will use a subset of the trigger set as the input to the DNN and observe the output. There will be a positive detection if certain probability requirements are met.

A successful watermarking method has to meet several criteria regarding its effectiveness, fidelity, false positive rate, and robustness. A detailed description of the criterion is presented in Table \ref{tab:criteria}. First, the effectiveness criterion states that a watermark has to ensure successful and consistent detection. The fidelity criterion states that watermarking cannot have a significant negative impact on the regular functionality of the model. False positive rate and robustness will be discussed separately in the following subsections.

\subsection{Proof of Ownership}
A watermarking method's ability to prove ownership mainly relies on its low false positive rate. Suppose that Alice decides that a watermark detection is positive if there are at most $\Delta$ misclassifications among $N$ trigger images. Then the probability of the detection can be calculated as follows \cite{GuoP18}\cite{abs-1804-00750}, assuming independence between the classification of each trigger image. $\rho$ represents the accuracy of the model on the trigger set.

\begin{equation}\label{eq:prob}
\Pr=\sum_{\delta=0}^\Delta {{N}\choose{\delta}}\rho^{N-\delta}(1 - \rho)^\delta
\end{equation}

The ownership is established based on the fact that $\Pr$ is disproportionally small for a non-watermarked neural network. If a watermarking method incurs a high false positive rate (a high $\rho$ for non-watermarked models), then $\Pr$ is no longer small and that the proof of ownership will be inconclusive at best.

\subsection{Threat Model}
In this subsection, we introduce our notion of robustness by defining our threat model. We assume that Bob has white-box access to the model, but does not have access to the training set. Instead, Bob has access to some proprietary test data (i.e. a subset of test set). We argue that proprietary data is one of the most important competitive advantages of Alice, and an IP pirate Bob by no means should have access to it. Otherwise, with the training data and the model, he might as well train a new model on his own, especially when he has the ability to carry out sophisticated attacks such as fine-tuning.  On the other hand, it is a reasonable assumption that the attacker may have white-box access to the model architecture and parameters. In the case of cloud ML service, Bob can be a malicious service platform. In the case of software ML libraries, Bob can be a hacker. In both cases, Bob would have the full white-box access to the model. We also assume that Alice only has black-box access to the model. In addition, Alice will have direct access to input to the model. There are no preprocessing stages between Alice's input and the input of the model.

With some test data and the model, Bob may fine-tune the model to produce a slightly different version of it. That is called the \textit{fine-tune} attack. After the fine-tune attack, Alice's watermark should still exist. Some researchers also discussed overwrite attacks, where Bob tries to embed his own watermark using the same procedure on Alice's model. It is indeed a very reasonable attack scenario. In our experiments, we found that embedding a new watermark using Bob's limited amount of data would adversely affect the model's performance, rendering the model much less valuable. Thus we rule out the possibility of Bob carrying out overwrite attacks.

\subsection{Problem Definition}
\label{defnition}

A DNN for classification is a function $f: \mathbb{R}^d \rightarrow [0, 1]^L $. Given an input $X \in \mathbb{R}^d$, it is desired that the function classifies it correctly to its label $y$, $f(X) = y$.

A pattern $P \in \mathbb{R}^d$ has the same dimensions as $X$, but is much more sparse. In its image form, $P$'s non-zero entries can be considered as a set of $K$ pixels with explicitly designed values and coordinates $\{v_k, c_k^x, c_k^y\} ^K$. $P$ is tightly coupled with the identity of the model owner. And the absolute and relative coordinates of the pixels may or may not contribute to $P$'s ability to carry information. The pattern can be embedded on \textit{any} input $X$ from the intended data distribution to convert it into a trigger input through a function $g(X, P)$ \footnote{For convenience, we sometimes write the function as $g(X_i, \{v_k, c_k^x, c_k^y\}^K)$. }. A watermarked DNN will be trained to classify $g(X, P)$ to $y_i' \neq y_i$. The fact that a DNN model classifies the trigger inputs disproportionally correctly can serve as a unique proof of the owner's identity. 

We consider two alternative approaches to create trigger patterns. In the method proposed by Guo \textit{et al.} (shown in Figure \ref{fig:other_3}), a color-coded key serves as the trigger pattern \cite{GuoP18}. They embed the pattern by offsetting the pixel values of the input, $g(X, P) = X + P$. Since the information is mainly ingrained in the pixel values, we consider the pixel locations ${c_k^x, c_k^y}$ to be flexible. We use \texttt{Key} throughout the paper to denote this trigger pattern. The second approach we consider is proposed by Zhang \textit{et al.}, and shown in Figure \ref{fig:other_2} \cite{ZhangGJWSHM18}. The information is obviously contained in the geometrical shape of the logo and the pixels have to remain in a relatively fixed to each other. Thus its location can be represented by its top left corner ${c^x, c^y}$. The author did not explicitly say how they embed the logo, but we interpret it as blending with the input, $g(X, P) =(1 - \alpha) X + \alpha P$. We denote the second type of trigger pattern as \texttt{Logo}.

Our main goal is to find the $P =\{c_k^x, c_k^y\}^K$  such that the probability of a non-watermarked DNN $f_0$ classifying a trigger input to its original labels is maximized. The main motivation behind the goal is to minimize false positive watermark detection. Empirically, given dataset $\mathcal{D}$, then the goal can be expressed as follows.

\begin{equation}\label{eq:pos}
\argmax_{\{c_k^x, c_k^y\}^K} |\{X | f_0(g(X, \{v_k, c_k^x, c_k^y\}^K)) = y, X, y \in \mathcal{D} \}|
\end{equation}

We have found that larger $v_k$ leads to more robust watermarking, although it leads to higher false positives. In the \texttt{Key}-related experiments, $v_k$ is given. But we can also integrate $v_k$ into the optimization landscape as follows. The \texttt{Logo}-related experiments use this objective function.

\begin{equation}\label{eq:strength}
\begin{split}
\argmax_{\{v_k, c_k^x, c_k^y\}^K} & |\{X | f_0(g(X, \{v_k, c_k^x, c_k^y\}^K)) = y, \\
& X, y \in \mathcal{D} \}| + 
  \delta \sum_k v_k
\end{split}
\end{equation}
$$$$

\section{Method}
\label{method}

In this section, we first provide a high level overview of why we chose the DE framework and how it works. Then we delve deeper to provide some algorithmic details that are crucial to the convergence of DE.

\subsection{Differential Evolution}
\begin{algorithm}
\caption{Differential Evolution}
\label{alg:de}
\textbf{Input}: dataset $\mathcal{D}$, non-watermarked DNN model $f_0$, populartion $N$, number of generations $G$\\
\textbf{Output}: best candidate $P$ after evolution
\begin{algorithmic}[1]
\State Randomly intialize, $i$th candidate $P_i=\{v_{ik}, c_{ik}^x, c_{ik}^y\}^K$
\For {generation $g=1, \ldots, G$}
	\For {each candidate $P_i$}
		\State Randomly pick $0 \leq j, k, l < N$ where $j \neq k \neq l$
		\State $P_i' \gets$ \textsc{evolve}($P_j$, $P_k$, $P_l$)
		\If {\textsc{fitness}($P_i', f_0, \mathcal{D}$) $>$ \textsc{fitness}($P_i, f_0, \mathcal{D}$)}
			\State $P_i \gets P_i'$
		\EndIf
	\EndFor
\EndFor
\State \Return $\argmax_{P_i} \textsc{fitness}(P_i, f_0, \mathcal{D})$
\end{algorithmic}
\end{algorithm}

To find the pattern $P$, the first methods that came up to us were the gradient-based methods commonly used for finding adversarial samples \cite{GoodfellowSS14} \cite{PapernotMJFCS16} \cite{Carlini017}. A key difference between our problem and theirs is that our pattern $P$ is universal. Therefore, finding the gradient of individual inputs hardly helps our situation. The family of evolutionary algorithms are among the most prominent non-gradient-based optimization methods. We initially relied on the generic evolutionary algorithm (EA), but we were unable to find a reasonable set of parameters to make the algorithm converge. That is when differential evolution (DE) presented itself as an alternative.

DE is a metaheuristic search algorithm that optimizes a given objective by evolving a population of candidates in parallel \cite{StornP97}. It follows the concept of generic EAs where a population of candidates evolves, and the candidates that are fittest will survive in each iteration. However, DE is simpler and it is known to facilitate faster convergence to the global optimum. Instead of using mutations and crossover between two parents, candidates DE evolve over a triplet. A new candidate is created by adding a weighted difference between two candidates to the third. 
\begin{equation}\label{eq:differential}
P=P_0 + F \times (P_1 - P_2)
\end{equation}

Algorithm \ref{alg:de} presents the high-level procedure of using DE to solve our problem. The main idea is to generate 1) new pixel coordinates 2) pixel values of $P$ using the differential variation operation described in Equation \ref{eq:differential}. The \textsc{fitness} function can be either of the two objective functions described in Section \ref{defnition}. The \textsc{evolve} function, on the other hand, is more complex. We describe more details of the function in the next subsection. 

\subsection{Optimizations for DE}

\begin{algorithm}[ht]
\caption{Evolve with Closest Triplet}
\label{alg:closest}
\textbf{Input}: 3 candidates, each containing $K$ pixel coordinates: $\{c_{1k}^x, c_{1k}^y\}^K$, $\{c_{2k}^x, c_{2k}^y\}^K$, $\{c_{3k}^x, c_{3k}^y\}^K$, differential weight $F$ \\
\textbf{Output}: pixel coordinates of a new candidate, $\{c_{k}^x, c_{k}^y\}^K$

\begin{algorithmic}[1]
\Procedure {Pair}{$\{c_{1k}^x, c_{1k}^y\}^K$, $\{c_{2k}^x, c_{2k}^y\}^K$}
	\State Initialize $heap$
	\For {each $\{c_{1i}^x, c_{1i}^y\}$} \Comment{Calculate pairwise distances}
		\For {each $\{c_{2j}^x, c_{2j}^y\}$}
			\State Push distance($\{c_{1i}^x, c_{1i}^y\}$, $\{c_{2j}^x, c_{2j}^y\}$) in $heap$
		\EndFor
	\EndFor
	\While {$heap$ is not empty} \Comment{Pair by distance}
		\State distance, $\{c_{1i}^x, c_{1i}^y\}$, $\{c_{2j}^x, c_{2j}^y\}$ $\gets$ Pop from $heap$
		\If {Neither $\{c_{1i}^x, c_{1i}^y\}$ or $\{c_{2j}^x, c_{2j}^y\}$ is paired}
			\State Pair ($\{c_{1i}^x, c_{1i}^y\}$ with $\{c_{2j}^x, c_{2j}^y\}$)
		\EndIf
	\EndWhile
\EndProcedure

\State \textsc{Pair}($\{c_{1k}^x, c_{1k}^y\}^K$, $\{c_{2k}^x, c_{2k}^y\}^K$)
\State \textsc{Pair}($\{c_{1k}^x, c_{1k}^y\}^K$, $\{c_{3k}^x, c_{3k}^y\}^K$)
\For {each $\{c_{2i}^x, c_{2i}^y\}^K$ $\{c_{3j}^x, c_{3j}^y\}^K$ paired with $\{c_{1k}^x, c_{1k}^y\}^K$ }
	\State $c_k^x \gets c_{1k}^x + F \times ( c_{2i}^x -  c_{3j}^x)$
	\State $c_k^y \gets c_{1k}^y + F \times ( c_{2i}^y -  c_{3j}^y)$ 	
\EndFor
\State \Return $\{c_{k}^x, c_{k}^y\}^K$
\end{algorithmic}
\end{algorithm}

We use two different variants of \textsc{evolve} functions for the two existing trigger patterns, \texttt{Key} and \texttt{Logo}. For \texttt{Logo}, the \textsc{evolve} function is straightforward. We use the top left pixel of the logo as the anchor, and each candidate can be represented by a simple triplet $(v, c^x, c^y)$. We use DE to evolve and select the location of the logo as well as its pixel values. We use a different approach for \texttt{Key}. Since pixel locations are al flexible, the candidate will be an array of $K$ tuples $\{c_k^x, c_k^y\}^K$. When we evolve using three candidates, each with $K$ pixels, which pixels should pair up and evolve becomes an important question. If pixels are randomly paired up, it is likely that the pixels will engage in a Brownian motion-like movement across different generations. Consequently, as we empirical show later, the evolution will not converge. If our goal is to evolve the pixels into optimal locations, then it makes sense to induce the evolution in such a way that pixels nearest to an optimal location will move toward that location. To that end, we propose an algorithm to pair closest pixels together to evolve. The most efficient implementation is to store all pairwise distances in heaps and always pair available pixels with the smallest distances. Algorithm \ref{alg:closest} describes implementation in more details. The time complexity of the algorithm is $O(K^2 log (K)$, where $K$ is the number of pixels. 

\begin{figure}[ht]
    \centering
    \includegraphics[width=2in]{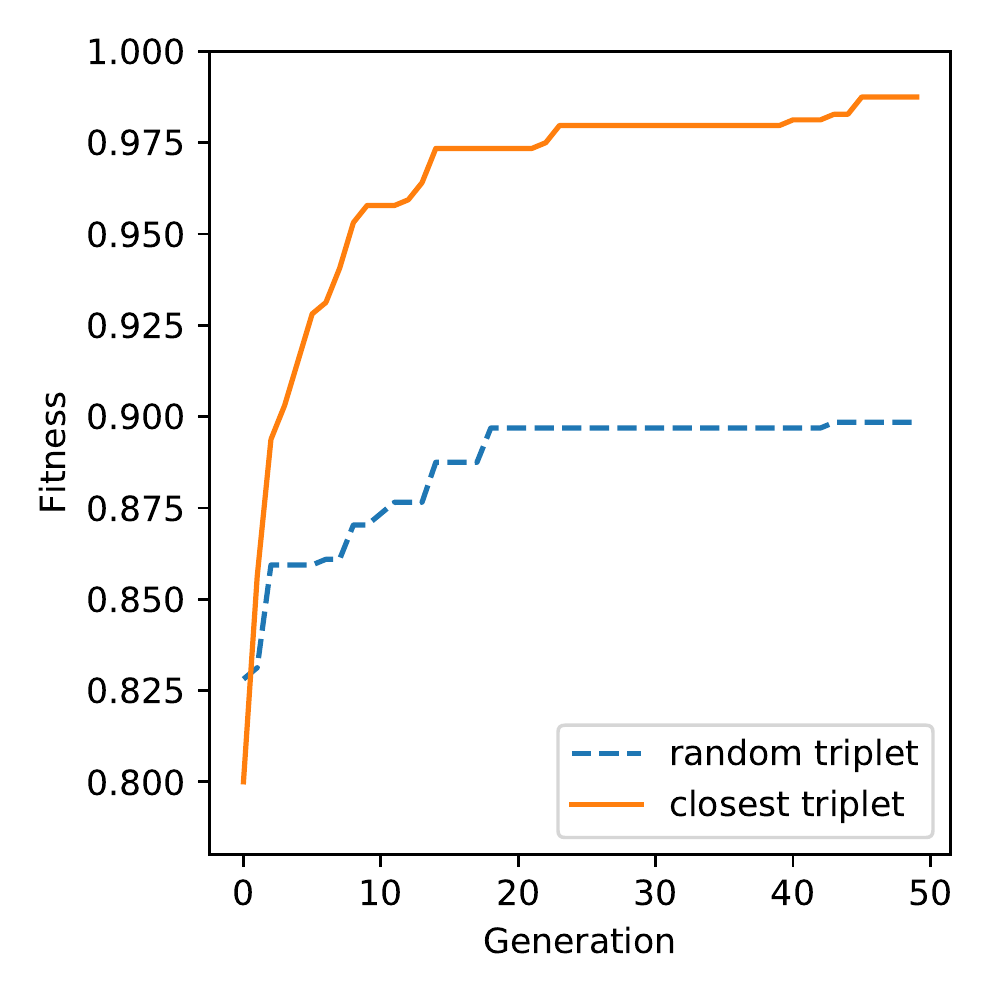}
    \caption{Best fitness of the population across the generations. }
      \label{fig:evolution}
\end{figure}

Figure \ref{fig:evolution} shows the fitness of the best candidate over the different generations. The \textsc{fitness} function is simply the accuracy of the subset. The proposed method, closest triplet evolve function, converged much faster than the evolve function where pixels are randomly paired together. In fact in the latter case, the fitness plateau at around 0.89 and it is unclear whether it will converge at all.

\section{Evaluation}
\label{evaluation}
In this section, we report the performance evaulations of our method. We first describe implementation details of the watermarking procedure and the DE algorithm. Then we evaluate the effectiveness, fidelity, false positive rate and robustness of the watermark in following subsections. Since neither \texttt{Logo} nor \texttt{Key} is an original idea from this paper, we omit many repetitive experiments for brevity. The key is to demonstrate the ability of our DE algorithm to reduce false positive while maintaining the robustness of the watermark. It is worth noting that \textit{all} of our trigger sets are built from test data that has not been used during training.

\subsection{Differential Evolution}

\begin{figure}[ht]
    \centering
    \subfigure[]{
        \includegraphics[width=1.1in]{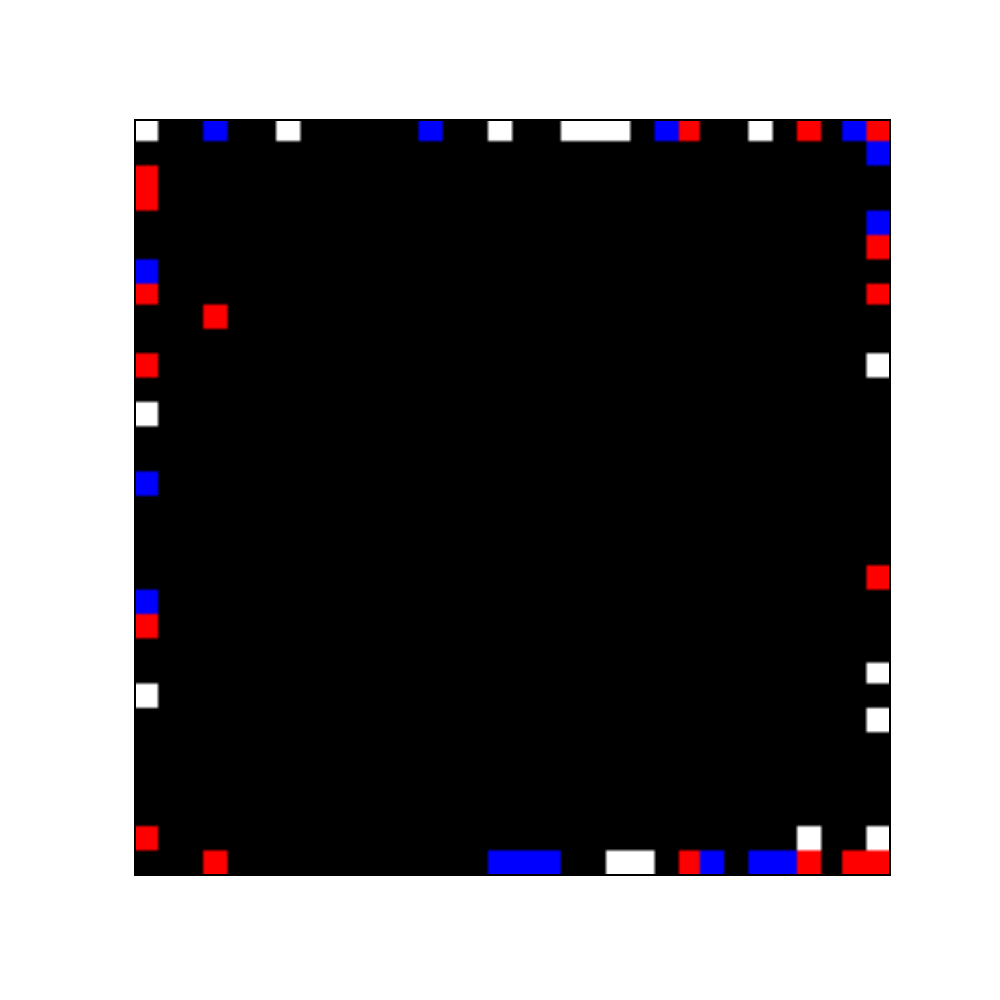}
        \label{fig:wm_key_cifar10}
    } 
    \hspace{-1.5em}
    \subfigure[]{
        \includegraphics[width=1.1in]{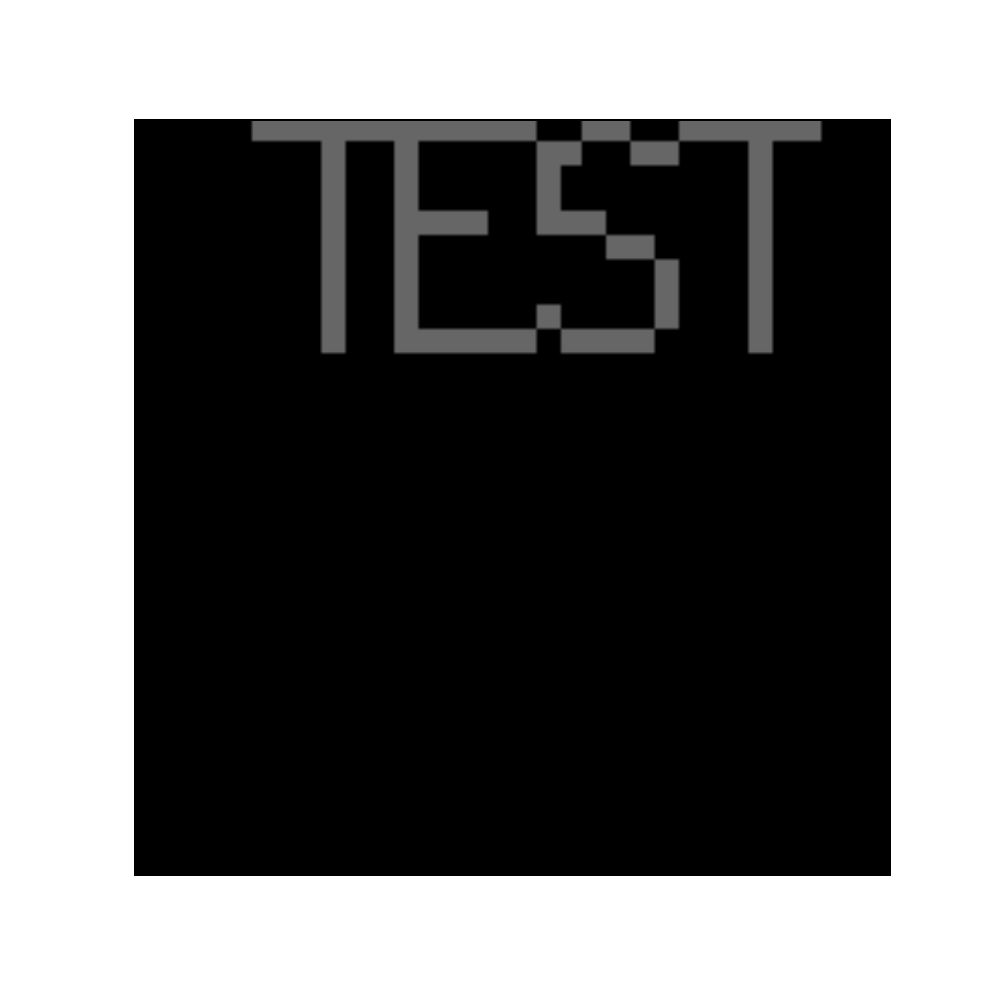}
        \label{fig:wm_logo_cifar10}
    }
    \hspace{-1.5em}    
    \subfigure[]{
        \includegraphics[width=1.1in]{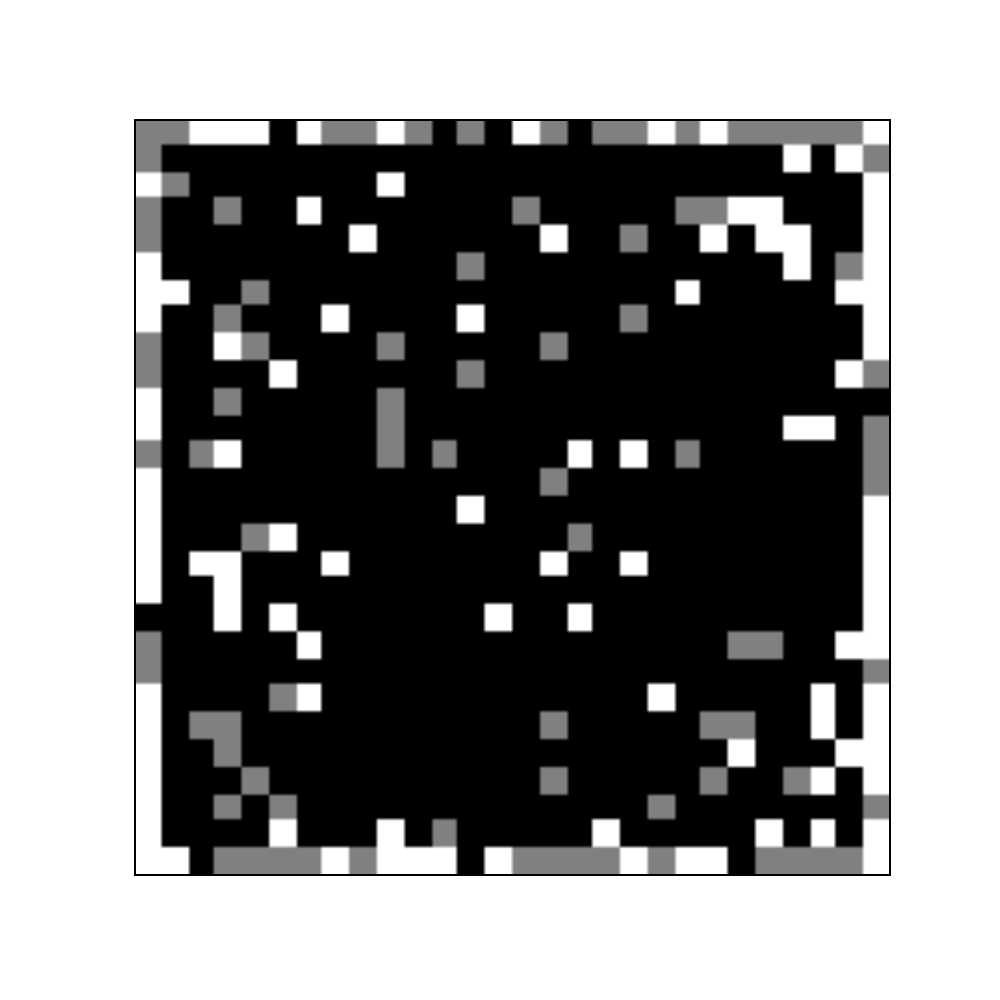}
        \label{fig:wm_key_mnist}
    } \\
    \vspace{-1em}
    \subfigure[]{
        \includegraphics[width=1.1in]{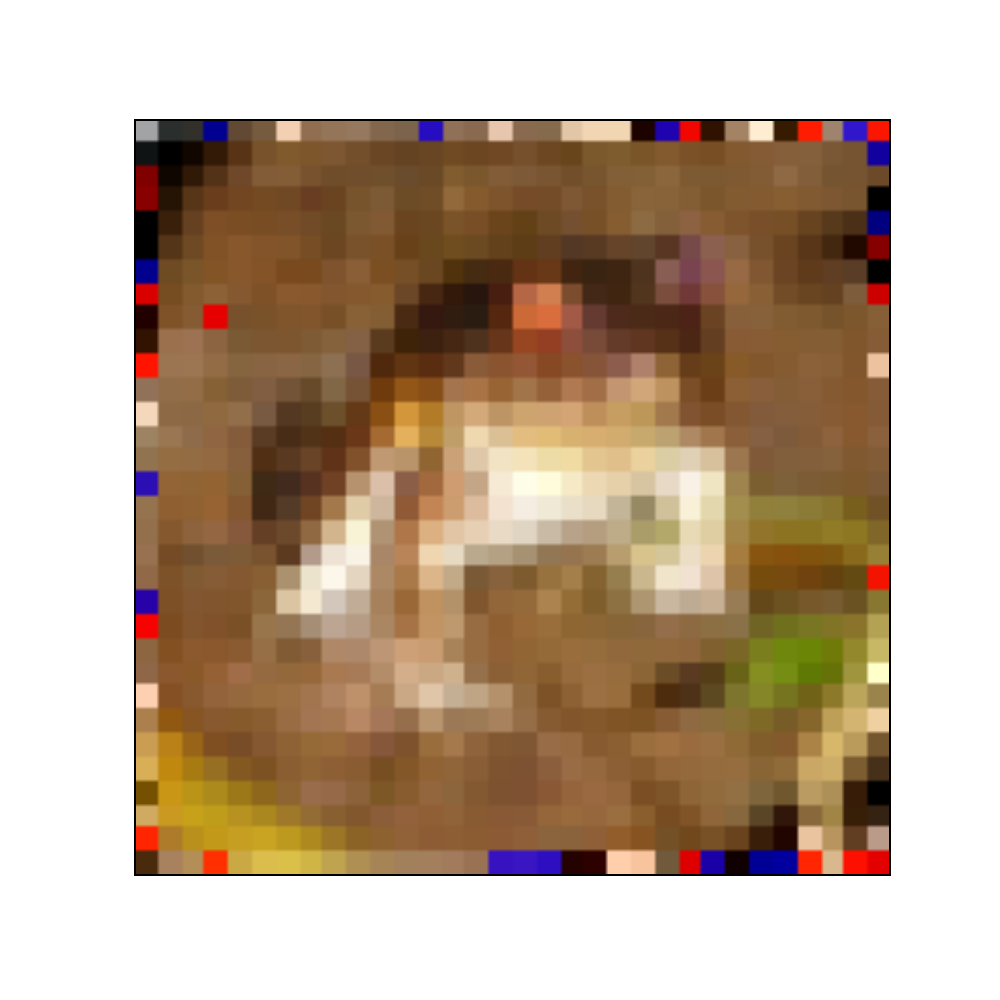}
        \label{fig:data_key_cifar10}
    } 
    \hspace{-1.5em}
    \subfigure[]{
        \includegraphics[width=1.1in]{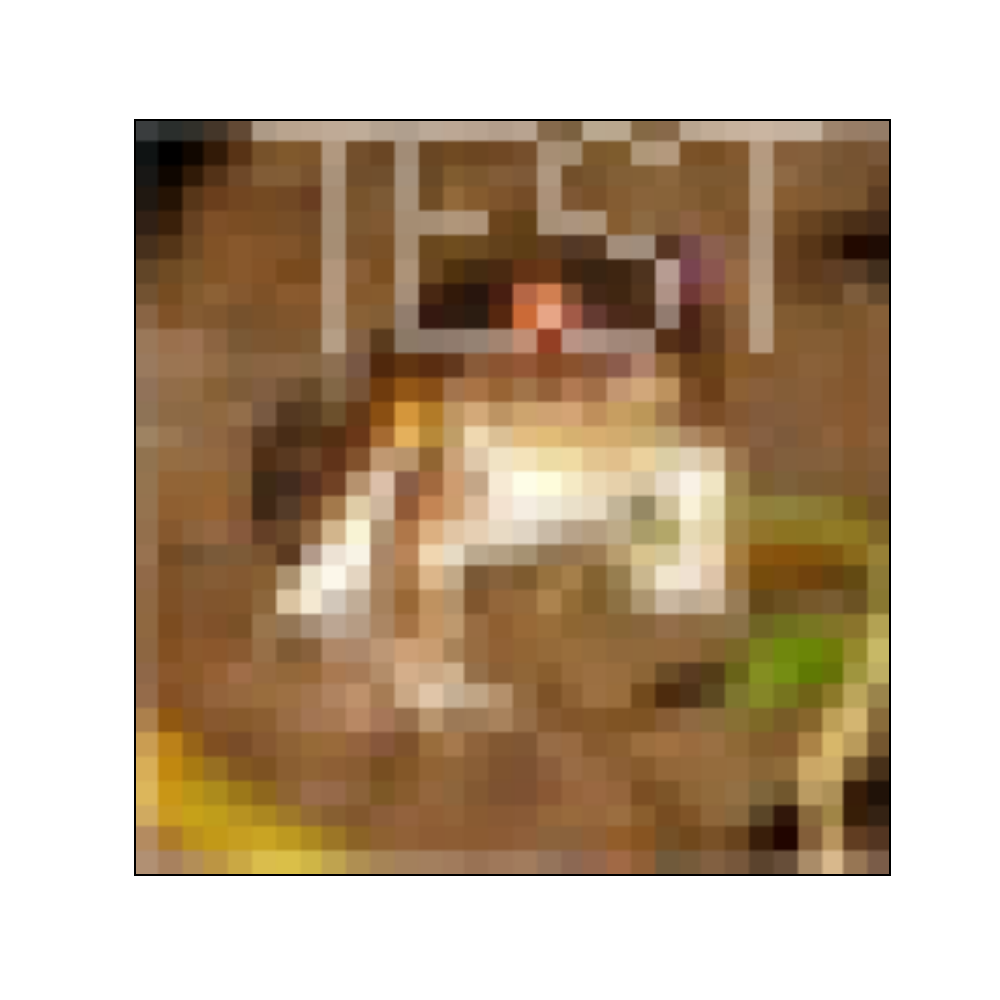}
        \label{fig:data_logo_cifar10}
    }
    \hspace{-1.5em}    
    \subfigure[]{
        \includegraphics[width=1.1in]{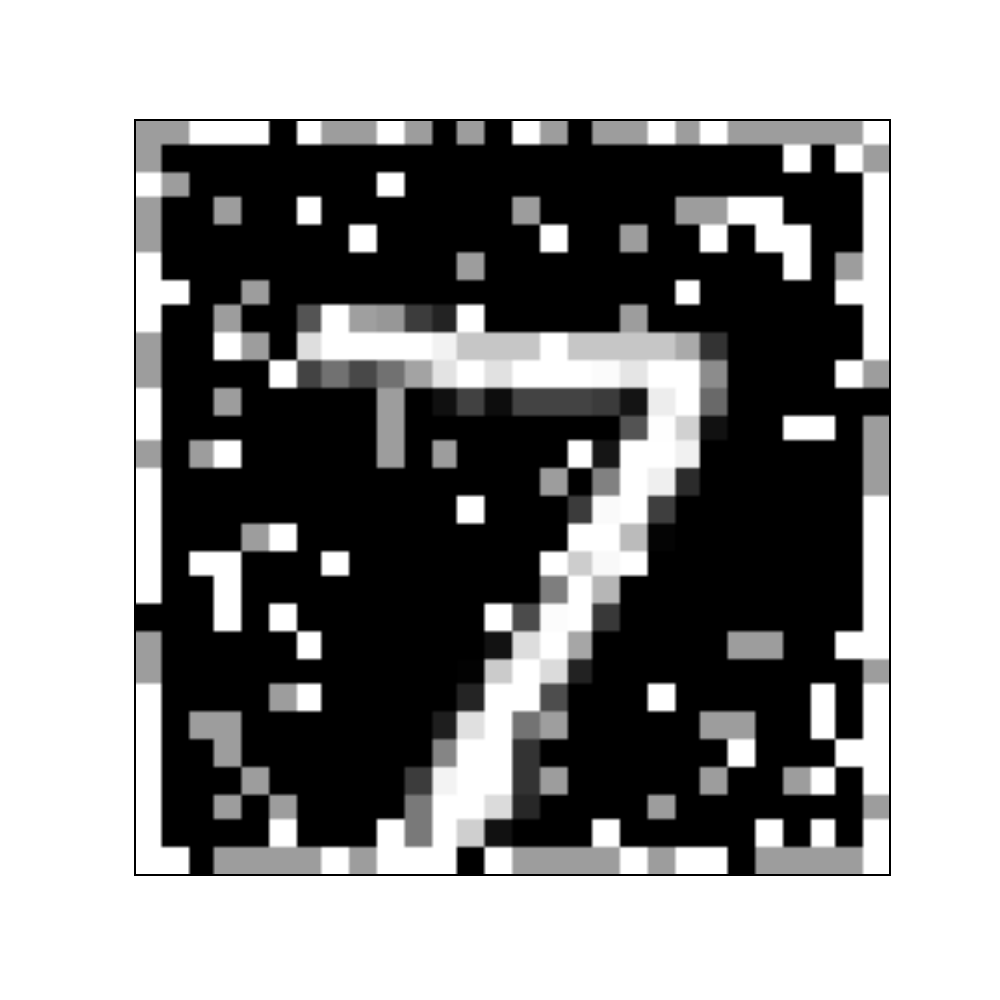}
        \label{fig:data_key_mnist}
    } \\    
    \caption{Output of our DE algorithm. (a) The \texttt{Key} trigger pattern on CIFAR-10, (b) the \texttt{Logo} trigger pattern on CIFAR-10, (c) the \texttt{Key} trigger pattern on MNIST, (d) a sample using the trigger pattern in (a), (e) a sample using the trigger pattern in (b), (f) a sample using the trigger pattern in (c).}
    \label{fig:our_watermark}
\end{figure}

We applied our DE-based approach on both \texttt{Logo} and \texttt{Key} trigger pattern generation. We use Equation \ref{eq:strength} as the fitness function for \texttt{Logo}, where both pixel locations and values are optimized. The weight $\beta$ is set such that a maximum $v$ constitutes 5\% of the fitness score. With the \texttt{Key} pattern, we only optimize the pixel locations. In the fitness functions of both DE algorithms, we evaluate the accuracy of candidates on a random set of 640 training images.  The model and dataset are described in the next subsection.

We carried out experiments on both the MNIST dataset and the CIFAR-10 dataset, and trigger patterns and trigger set images output by the DE algrithm are shown in Figure \ref{fig:our_watermark}. To create \texttt{Logo} patterns on the CIFAR-10 dataset, we replicated the experiments by Zhang \textit{et al.} \cite{ZhangGJWSHM18}. Surprisingly, the optimal location to put the logo isn't at one of the 4 corners as one would intuitively think. In DE, we set $v_k=255$ searched the coefficient for blending $\alpha$ instead, which yielded an optimal value of 0.4019.

Like Guo \textit{et al.} \cite{GuoP18}, we encoded the message in the pixel values of the \texttt{Key} trigger pattern. The CIFAR-10 variant includes 64 pixels and encodes 128 bits of information. Every pixel in the RGB color space with $v_k=\pm100$ encodes 2 bits, and the message can be decoded by reading the pixels from left to right, top to bottom. The MNIST variant includes 192 pixels and encodes 192-bit information, with $v_k=100, 200$ to encode 0 and 1 respectively. In both cases, pixels in the \texttt{Key} pattern gravitate towards the edge of the images. Clearly, the DE algorithm is rewarding pixel locations that do not overlap with the objects, which tend to occupy the center of the image. The new pixel locations form patterns in a way that has minimal impacts on the classification of an object. Because of that, even we added patterns with large pixel values, the resulting images still didn't trigger regular models.

To test the capacity of our algorithm, we intentionally used patterns that have a lot more complexity in our experiments on the MNIST dataset. Images in MNIST dataset has a lot more empty space to take advantage of, while pixels blindly accumulate at the edge may cause misclassification. Through our algorithm, the probability $\Pr(f_0(g(X_i, P))=y_i)$ increased from as low as 83.30\% (during random initialization) to 99.27\%. The probability is measured over the entire trigger set, and the classification accuracy on the regular test set with the exact same images is 99.46. It clearly shows the algorithms ability of learning to reduce false positive detections of the watermark. To test our DE algorithm's ability to converge, we repeated the MNIST/CIFAR10 "key" on 8 different parameter sets (number of pixels in the pattern, etc.), 5 experiments per set. Eat set all converged to solutions that produce extremely similar fitness scores, with an average standard deviation of 0.0060.

\subsection{Effectiveness and Fidelity}
\begin{table}
\centering
\caption{Classification accuracy of watermarked models on test sets and trigger tests. The trigger sets are derived from the test sets using the \texttt{Key} / \texttt{Logo} method.}
\label{tab:effectiveness}
\begin{tabular}{c|c|c|c}
\hline
 & Dataset & \thead{Regular} & \thead{Watermarked} \\ \hline
\multirow{4}{*}{\thead{CIFAR-10 \\ResNet-18}} &Test & 94.13 & 93.49 \\
&Trigger (\texttt{Key}) & 1.08 & 93.67 \\
&Test & 94.13 & 93.76 \\
&Trigger (\texttt{Logo}) & 1.27 &93.49 \\ \hline
\multirow{2}{*}{\thead{MNIST \\4 CONV-2 FC}} &Test & 99.46 & 99.38 \\
&Trigger (\texttt{Key}) & 0.25 & 99.26 \\  \hline

\end{tabular}
\end{table}

It has been demonstrated in all previous works that DNNs can be trained to successfully recognize the triggers sets. In addition, Adi \textit{et al.} also showed the significance of start training from scratch in creating a robust watermark \cite{AdiBCPK18}. We followed the same procedure. Table \ref{tab:effectiveness} shows the classification accuracy of both non-watermarked and watermarked models on the regular test set and the trigger set. 

In light of the fidelity criterion, the classification accuracy of the watermarked model on the regular test set is sligtly lower compared to the regular non-watermarked model. It is expected as watermarking makes the classification problem much harder. In light of the effectiveness criterion, the ability of the watermarked model to recognize the trigger set is as good as its ability to classify regular images.

\subsection{False Positives}

\begin{figure}[ht]
    \centering
    \includegraphics[width=3in]{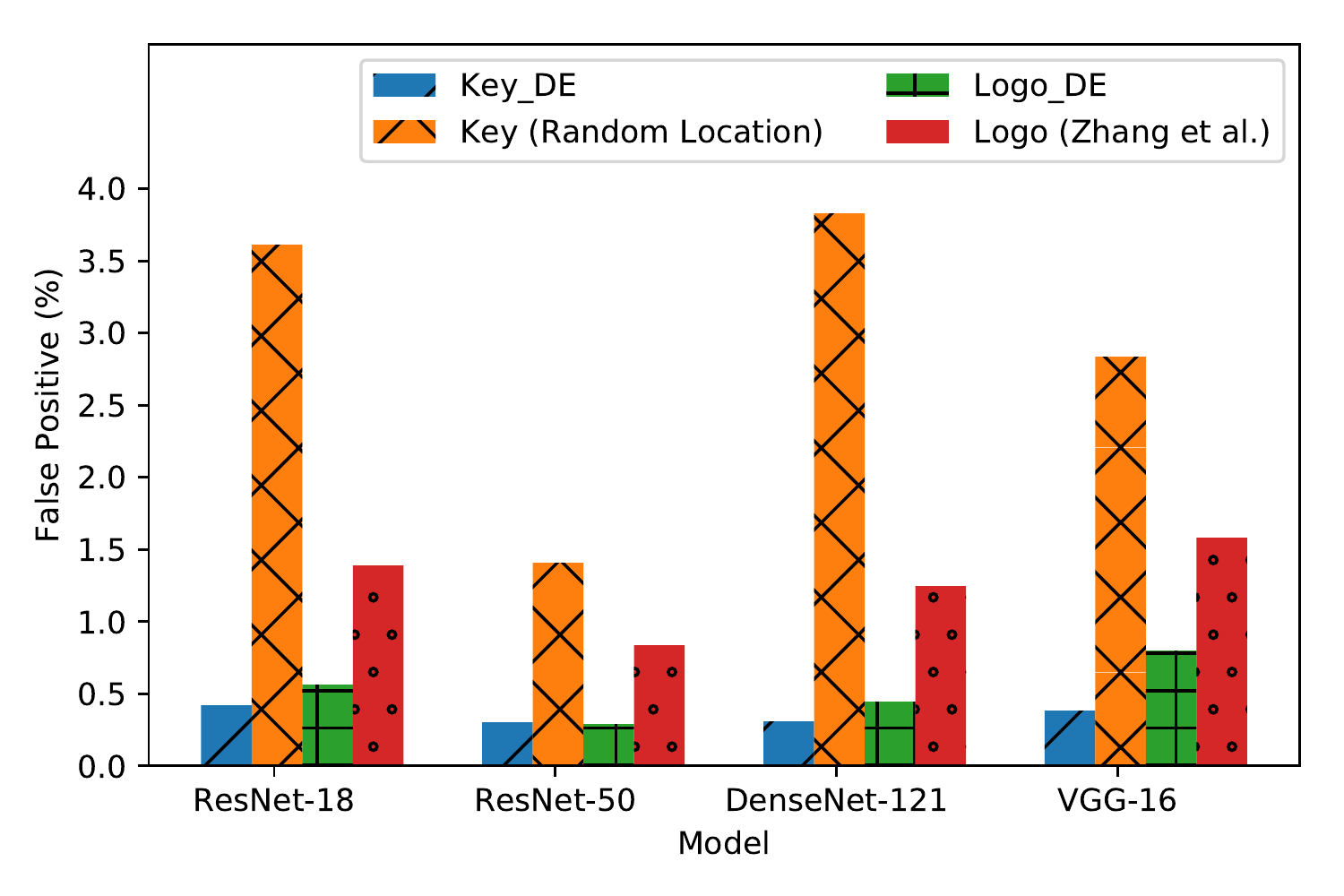}
    \caption{False positive rates of different CIFAR-10 trigger sets. It is the probability of a non-watermarked DNN getting falsely triggered. The lower the false positive rate the better.}
    \label{fig:false_positive}
\end{figure}

Figure \ref{fig:false_positive} shows the false positive rate of different trigger patterns. The false postive rate here is measured by the probability that a non-watermarked model classifies a trigger image into its re-assigned class $\Pr(f_0(g(X_i, P)=y_i')$. We used four different non-watermarked DNN trained on the regular CIFAR-10 dataset: ResNet-18, ResNet-50, DenseNet-121, VGG-16. The fitness function in DE used to obtain the trigger pattern only involves the ResNet-18 model. The results show that the what our DE learns from one model generalizes well to other models as well. We tested the generalizability further using the same pattern on 5 newly trained VGG-13s, and obtained a 95\% confidence interval of $1.15\% \pm 0.06\%$. 

We used two baselines for comparison. To compare with the DE-based \texttt{Key} pattern, we used a \texttt{Key} pattern with random $\{x_k, y_k\}^K$ but the same $v_k$. We see drastic improvements with up to 10$\times$ reduction in the false positive rate. Note that the trigger pattern proposed by  Guo \textit{et al.} is also based on random location \cite{GuoP18}. But they explicitly selected $v_k$ such that the pattern is imperceptible, resulting in a lower false positive rate. But as we see later, they achieved that at the cost of robustness. To compare with the DE-based \texttt{Logo}, we use \texttt{Logo} trigger pattern used Zhang \textit{et al.} \cite{ZhangGJWSHM18}. We achieve about 2$\times$ improvement in false positive rate. Putting it in the perspective of Equation \ref{eq:prob}, even 2$\times$ translates to over $25000\times$ lower probability ($\Delta=5, N=20$). 

\subsection{Robustness}
We measure the robustness of the watermarking methods through their resistance against fine-tune attacks. Table \ref{tab:robustness} reports trigger set classification accuracy loss after we fine-tuned a watermarked model. Unlike some of the earlier approaches that based their attack on the training set, we used 1000 test images and applied various data augmentation techniques. The model watermarked using the original \texttt{key} method suffered a significant drop in accuracy. The accuracy drop was almost entirely eliminated when we switch to our \texttt{key} method. Both of the \texttt{logo} method were resilient against the fine-tune attack.

Images superimposed with our pattern are sufficiently different from the normal input distribution. Because of that, a watermarked model's ability to recognize those patterns is largely orthogonal to its ability to classify objects and is, therefore, harder to remove during a fine-tune attack.

\subsection{Discussions}

Gradient-free methods do exist in the world of adversarial learning, most notably in the subproblem of black-box attacks, where the attackers don't have access to the gradient information \cite{nguyen2015deep}\cite{abs-1710-08864}\cite{IlyasEAL18}\cite{abs-1805-11090}. Those methods again focus on individual samples and are essentially solving a different problem than ours. It is worth noting that the work from Moosavi-Dezfooli \textit{et al.} aims at creating universal adversarial perturbations \cite{Moosavi-Dezfooli17}. Their proposal to reduce the search space to a subset of the input provided invaluable insights.

Due to the limited scope of this paper, the parameter $v_k$ isn't systematically studied. It is more a heuristic and manually selected in many situations. It would be valuable to study how it systematically affects the robustness of the watermarking methods.

%\begin{figure}
%    \centering
%    \includegraphics[width=3in]{figures/finetune.pdf}
%    \caption{Classification accuracy loss after fine-tune attack.}
%    \label{fig:finetune}
%\end{figure}

\begin{table}
\centering
\caption{Classification accuracy of watermarked models on corresponding CIFAR-10 trigger sets after fine-tune attacks. The less the accuracy loss, the more robust the method is.}
\label{tab:robustness}
\begin{tabular}{c|c}
\hline
Method & Accuracy loss (\%) \\ \hline
\texttt{Key}, DE & -0.29 \\
\texttt{Key}, \cite{GuoP18} & -73.97 \\ \hline
\texttt{Logo}, DE & -1.01 \\
\texttt{Logo}, \cite{ZhangGJWSHM18} & -0.97\\ \hline
\end{tabular}
\end{table}

\section{Conclusion}

Black-box DNN watermarking has emerged as a viable solution to IP protection in the context of MLaaS. Adding owner identity-based trigger patterns to natural input images is a popular method to create effective trigger sets that establish strong ownership proofs. In this paper, we propose a novel differential evolution-based framework to optimize the generation of such trigger patterns. Compared to the prior art, our method demonstrates significant improvement in false positive rate and robustness in experiments on popular models and datasets.

\bibliographystyle{ieeetr}
\bibliography{bib}

\begin{thebibliography}{10}

\bibitem{ai-investment}
L.~Chapman, ``Vcs plowed a record \$ 9.3 billion into ai startups last year.''
  \url{https://www.bloomberg.com/news/articles/2019-01-08/vcs-plowed-a-record-9-3-billion-into-ai-startups-last-year},
  2019.

\bibitem{UchidaNSS17}
Y.~Uchida, Y.~Nagai, S.~Sakazawa, and S.~Satoh, ``Embedding watermarks into
  deep neural networks,'' in {\em Proceedings of {ACM} International Conference
  on Multimedia Retrieval}, pp.~269--277, 2017.

\bibitem{abs-1711-01894}
E.~L. Merrer, P.~Perez, and G.~Tr{\'{e}}dan, ``Adversarial frontier stitching
  for remote neural network watermarking,'' {\em CoRR}, vol.~abs/1711.01894,
  2017.

\bibitem{AdiBCPK18}
Y.~Adi, C.~Baum, M.~Ciss{\'{e}}, B.~Pinkas, and J.~Keshet, ``Turning your
  weakness into a strength: Watermarking deep neural networks by backdooring,''
  in {\em 27th {USENIX} Security Symposium}, pp.~1615--1631, 2018.

\bibitem{abs-1804-00750}
B.~D. Rohani, H.~Chen, and F.~Koushanfar, ``Deepsigns: A generic watermarking
  framework for ip protection of deep learning models,'' {\em CoRR},
  vol.~abs/1804.00750, 2018.

\bibitem{ZhangGJWSHM18}
J.~Zhang, Z.~Gu, J.~Jang, H.~Wu, M.~P. Stoecklin, H.~Huang, and I.~Molloy,
  ``Protecting intellectual property of deep neural networks with
  watermarking,'' in {\em Proceedings of the 2018 on Asia Conference on
  Computer and Communications Security}, pp.~159--172, 2018.

\bibitem{GuoP18}
J.~Guo and M.~Potkonjak, ``Watermarking deep neural networks for embedded
  systems,'' in {\em Proceedings of the International Conference on
  Computer-Aided Design}, p.~133, 2018.

\bibitem{CollbergT99}
C.~S. Collberg and C.~D. Thomborson, ``Software watermarking: Models and
  dynamic embeddings,'' in {\em Proceedings of the 26th {ACM} {SIGPLAN-SIGACT}
  Symposium on Principles of Programming Languages}, pp.~311--324, 1999.

\bibitem{KahngLMMMPTWW98}
A.~B. Kahng, J.~Lach, W.~H. Mangione{-}Smith, S.~Mantik, I.~L. Markov,
  M.~Potkonjak, P.~Tucker, H.~Wang, and G.~Wolfe, ``Watermarking techniques for
  intellectual property protection,'' in {\em Proceedings of the 35th
  Conference on Design Automation}, {DAC}, pp.~776--781, 1998.

\bibitem{GoodfellowSS14}
I.~J. Goodfellow, J.~Shlens, and C.~Szegedy, ``Explaining and harnessing
  adversarial examples,'' {\em CoRR}, vol.~abs/1412.6572, 2014.

\bibitem{PapernotMJFCS16}
N.~Papernot, P.~D. McDaniel, S.~Jha, M.~Fredrikson, Z.~B. Celik, and A.~Swami,
  ``The limitations of deep learning in adversarial settings,'' in {\em {IEEE}
  European Symposium on Security and Privacy}, pp.~372--387, 2016.

\bibitem{Carlini017}
N.~Carlini and D.~A. Wagner, ``Towards evaluating the robustness of neural
  networks,'' in {\em 2017 {IEEE} Symposium on Security and Privacy},
  pp.~39--57, 2017.

\bibitem{StornP97}
R.~Storn and K.~V. Price, ``Differential evolution - {A} simple and efficient
  heuristic for global optimization over continuous spaces,'' {\em J. Global
  Optimization}, vol.~11, no.~4, pp.~341--359, 1997.

\bibitem{nguyen2015deep}
A.~Nguyen, J.~Yosinski, and J.~Clune, ``Deep neural networks are easily fooled:
  High confidence predictions for unrecognizable images,'' in {\em {IEEE}
  Conference on Computer Vision and Pattern Recognition}, pp.~427--436, 2015.

\bibitem{abs-1710-08864}
J.~Su, D.~V. Vargas, and K.~Sakurai, ``One pixel attack for fooling deep neural
  networks,'' {\em CoRR}, vol.~abs/1710.08864, 2017.

\bibitem{IlyasEAL18}
A.~Ilyas, L.~Engstrom, A.~Athalye, and J.~Lin, ``Black-box adversarial attacks
  with limited queries and information,'' in {\em Proceedings of the 35th
  International Conference on Machine Learning}, pp.~2142--2151, 2018.

\bibitem{abs-1805-11090}
M.~Alzantot, Y.~Sharma, S.~Chakraborty, and M.~B. Srivastava, ``Genattack:
  Practical black-box attacks with gradient-free optimization,'' {\em CoRR},
  vol.~abs/1805.11090, 2018.

\bibitem{Moosavi-Dezfooli17}
S.~Moosavi{-}Dezfooli, A.~Fawzi, O.~Fawzi, and P.~Frossard, ``Universal
  adversarial perturbations,'' in {\em 2017 {IEEE} Conference on Computer
  Vision and Pattern Recognition}, pp.~86--94, 2017.

\end{thebibliography}

\end{document}